\documentclass[letterpaper, 10 pt, conference]{ieeeconf}  
\usepackage{amsmath}
\usepackage{amsfonts}
\usepackage{graphicx}
\usepackage{booktabs}
\usepackage{multirow}
\usepackage{tabularx}
\usepackage{siunitx}
\usepackage{subfigure}
\usepackage{float}
\usepackage{amssymb}
\usepackage{mathtools}
\usepackage{url}
\usepackage{xcolor}
\usepackage{xpatch}
\usepackage{flushend}
\makeatletter
\definecolor{lightblue}{RGB}{54, 125, 189}
\let\oldcite\cite
\renewcommand{\cite}[1]{\textcolor{lightblue}{\oldcite{#1}}}
\let\oldref\ref
\renewcommand{\ref}[1]{\textcolor{lightblue}{\oldref{#1}}}
\makeatother

\IEEEoverridecommandlockouts                              

\title{\LARGE \bf
VisLanding: Monocular 3D Perception for UAV Safe Landing via Depth-Normal Synergy
}

\author{Zhuoyue Tan$^{1*}$, Boyong He$^{1*}$, Yuxiang Ji$^{1}$, Liaoni Wu$^{1,2\dagger}$
\thanks{$^{1}$Institute of Artificial Intelligence, Xiamen University}%
\thanks{$^{2}$School of Aerospace Engineering, Xiamen University}%
\thanks{$^{*}$These authors contributed equally}%
\thanks{$^{\dagger}$Corresponding author. Email: wuliaoni@xmu.edu.cn}%
}

\begin{document}

\maketitle
\thispagestyle{empty}
\pagestyle{empty}

\begin{abstract}
This paper presents \textbf{VisLanding}, a monocular 3D perception-based framework for safe UAV (Unmanned Aerial Vehicle) landing. Addressing the core challenge of autonomous UAV landing in complex and unknown environments, this study innovatively leverages the depth-normal synergy prediction capabilities of the Metric3D V2 model to construct an end-to-end safe landing zones (SLZ) estimation framework. By introducing a safe zone segmentation branch, we transform the landing zone estimation task into a binary semantic segmentation problem. The model is fine-tuned and annotated using the WildUAV dataset from a UAV perspective, while a cross-domain evaluation dataset is constructed to validate the model's robustness. Experimental results demonstrate that \textbf{VisLanding} significantly enhances the accuracy of safe zone identification through a depth-normal joint optimization mechanism, while retaining the zero-shot generalization advantages of Metric3D V2. The proposed method exhibits superior generalization and robustness in cross-domain testing compared to other approaches. Furthermore, it enables the estimation of landing zone area by integrating predicted depth and normal information, providing critical decision-making support for practical applications. 
\end{abstract}

\section{INTRODUCTION}
Nowadays, the application domains of Unmanned Aerial Vehicles (UAVs) have rapidly expanded, penetrating extensively into critical fields such as military, civilian, and commercial sectors~\cite{c1}. Leveraging their flexibility and adaptability, UAVs have demonstrated significant advantages in diverse tasks, including surveillance and monitoring, facility inspection, logistics delivery, and recreational applications. Among these, achieving safe and autonomous landing in complex and unknown environments stands out as one of the most challenging key technologies for UAVs, holding substantial application value in emergency response scenarios and situations with limited human intervention. Traditional landing methods primarily rely on predefined markers or specific landing points; however, these prerequisites are often difficult to satisfy in practical application scenarios. Therefore, the development of advanced algorithms and technologies to enable UAVs to autonomously identify safe landing zones has become a crucial research topic~\cite{c2}.
\begin{figure}[thpb]
  \centering
  \includegraphics[width=0.46\textwidth]{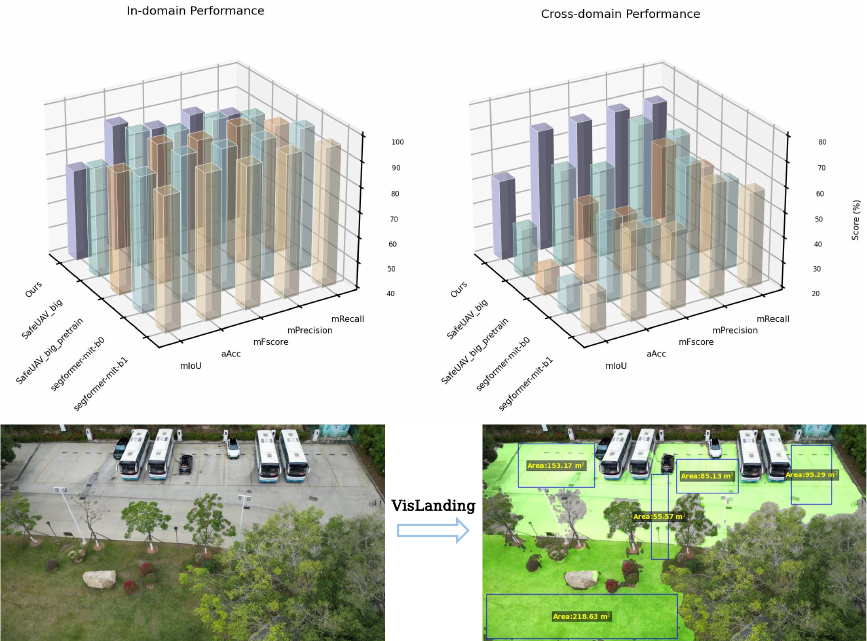}
  \caption{\textbf{Top:} Comparative performance of our method versus baseline approaches on both \textit{In-domain} and \textit{Cross-domain} test benchmarks for the SLZ estimation task. The proposed method demonstrates enhanced robustness and superior generalization capabilities across all key metrics. 
\textbf{Bottom:} Zero-shot testing results on real-world flight data. The green mask indicates pixels predicted as safe landing zones, blue bounding boxes highlight the top-5 landing candidate regions based on area calculation, and yellow annotations display the estimated areas of candidate zones. These results validate the practical applicability of our method in real-world scenarios.
}
  \label{fig3d}
\end{figure}
The core challenge for UAVs to achieve safe landing in unknown regions lies in accurately identifying suitable safe landing areas~\cite{c3}. Existing methods fall into two categories: multi-sensor fusion and pure vision-based algorithms. Multi-sensor approaches integrate data from RGB-D cameras, LiDAR, and IMUs for high-precision terrain reconstruction, but suffer from high hardware costs. Pure vision-based methods, while cost-effective, struggle with scale ambiguity and geometric reliability~\cite{c2}.

Fortunately, monocular 3D perception technology, represented by Metric3D V2, has recently achieved breakthrough progress. This technology, through deep learning methods, realizes joint optimization of depth prediction and normal estimation, achieving state-of-the-art (SOTA) performance across multiple benchmarks. Its exceptional zero-shot generalization capability establishes it as a foundational model in the field of monocular 3D perception~\cite{metric3d}. Depth information and surface normal information provide UAVs with critical cues regarding landing zone distance and terrain characteristics, offering new research directions for SLZ estimation based on pure vision algorithms.

Building upon the Metric3D V2 framework, we propose an end-to-end UAV SLZ estimation framework---\textbf{VisLanding}. Specifically, we introduce a new safe zone segmentation branch to the existing Metric3D V2 framework, leveraging its depth prediction and normal estimation capabilities as auxiliary features. The safe landing zone estimation task is transformed into a binary semantic segmentation problem, enabling the delineation of safe zones within monocular images. We first fine-tuned the model using the WildUAV dataset~\cite{wilduav} to enhance its monocular depth estimation capability from a UAV perspective. As one of the few high-quality real-world datasets for UAV Monocular Depth Estimate tasks, WildUAV effectively mitigates domain discrepancy issues associated with synthetic data. Furthermore, we manually annotated safe landing zone labels for the mapping set of the WildUAV dataset with depth annotations to train the model's safe zone estimation capability. In addition, we constructed an evaluation dataset using the Semantic Drone Dataset~\cite{semantic_drone} to test the model's cross-domain generalization and robustness. The main contributions of this work are summarized as follows:
\begin{itemize}
\item We innovatively employ the Metric3D V2 model to replace complex multi-sensor systems for acquiring critical 3D information required for SLZ identification. This approach not only fully leverages Metric3D V2's strengths in depth prediction and surface normal estimation but also significantly reduces system complexity and hardware costs.
\item We constructed a training dataset and a cross-domain evaluation dataset based on real-world scenarios, providing essential data support for research on SLZ estimation using monocular vision algorithms.
\item As demonstrated in Fig.~\ref{fig3d}, the proposed method demonstrates excellent performance with strong generalization and robustness, outperforming existing approaches. It combines post-processing to achieve direct estimation of the safe landing zone area, better aligning with practical application requirements, and providing an effective solution for safe landing of UAVs.
\end{itemize}

\section{RELATED WORK}

\subsection{Monocular 3D Perception}

Monocular 3D perception, as a fundamental task in computer vision, faces significant challenges in metric-scale reconstruction due to the inherent lack of depth information and scale ambiguity in single-view images. Recent research has demonstrated substantial application value in practical scenarios such as autonomous driving through cross-domain generalization strategies and unsupervised learning frameworks. ZoeDepth~\cite{c6} innovatively combines pre-trained relative depth models with metric-scale fine-tuning strategies, establishing a new paradigm for scale-sensitive depth estimation. UniDepth~\cite{c7} introduces a self-promptable camera module that decouples camera parameters from depth features through pseudo-spherical representation, enabling cross-domain metric reconstruction from single views. ZeroDepth~\cite{c8} builds upon the Perceiver IO architecture to construct a unified multi-domain framework, effectively addressing the challenge of absolute depth prediction in both indoor and outdoor scenes. Depth Anything V2~\cite{c9} leverages large-scale unlabeled data to establish a data-driven paradigm, significantly reducing generalization errors. Diffusion-based approaches, such as Marigold~\cite{c10} and DiffusionDepth~\cite{c11}, enhance detail reconstruction accuracy through probabilistic modeling mechanisms. Metric3D V2~\cite{metric3d} proposes the concept of a canonical camera space, effectively resolving scale ambiguity by unifying camera parameter representations. This approach achieves joint depth-normal estimation under zero-shot conditions. It attains state-of-the-art performance on multiple benchmarks, establishing itself as a foundational model, providing strong support for metric 3D reconstruction in open-world scenarios.

\subsection{Safe Landing Zones Estimation}

With the increasing demand for autonomous operations of UAVs in both civilian and military fields, the estimation of SLZ has garnered significant attention from both academia and industry. To address the technical challenges of identifying suitable landing areas in non-cooperative or dynamic environments, researchers have proposed various solutions from multiple dimensions. In recent years, thanks to the rapid development of deep learning technologies, computer vision has become one of the key technologies for achieving autonomous UAV landing. Early studies~\cite{c12,c13,c14,c15} have explored the use of deep neural networks for SLZ estimation. Building on this, SafeUAV~\cite{safeuav} innovatively combined depth estimation with SLZ estimation, proposing a specialized synthetic dataset for task training. This system can achieve depth estimation through RGB images and classify terrains into categories such as horizontal and vertical to distinguish SLZ, demonstrating good performance while ensuring real-time capabilities. Chen et al. (2022)~\cite{c17} developed an autonomous landing system based on a binocular LiDAR sensor system, which integrates a terrain understanding model to achieve simultaneous depth completion and semantic segmentation. This method enables UAVs to accurately infer the morphological features and semantic information of the terrain, thereby achieving high-precision SLZ estimation in complex environments, albeit with relatively high system complexity and hardware costs. Abdollahzadeh et al. (2022)~\cite{c18} proposed a depth regression model based on a semantic segmentation framework, which can generate continuous safety score maps, providing a more refined landing safety assessment compared to traditional binary classification methods. Additionally, Serrano and Bandala (2023)~\cite{c19} innovatively applied the YOLO (You Only Look Once)~\cite{c20} series of object detection algorithms to safe landing zones estimation tasks. In the latest research, Loera-Ponce et al. (2024)~\cite{c21} employed the advanced vision transformer network SegFormer~\cite{c22} to perform semantic segmentation on images captured by UAVs. By mapping segmentation categories to different risk levels for risk assessment, they provided an effective solution for SLZ estimation in emergency situations.

\begin{figure*}[thpb]
  \centering
  \includegraphics[width=0.98\textwidth]{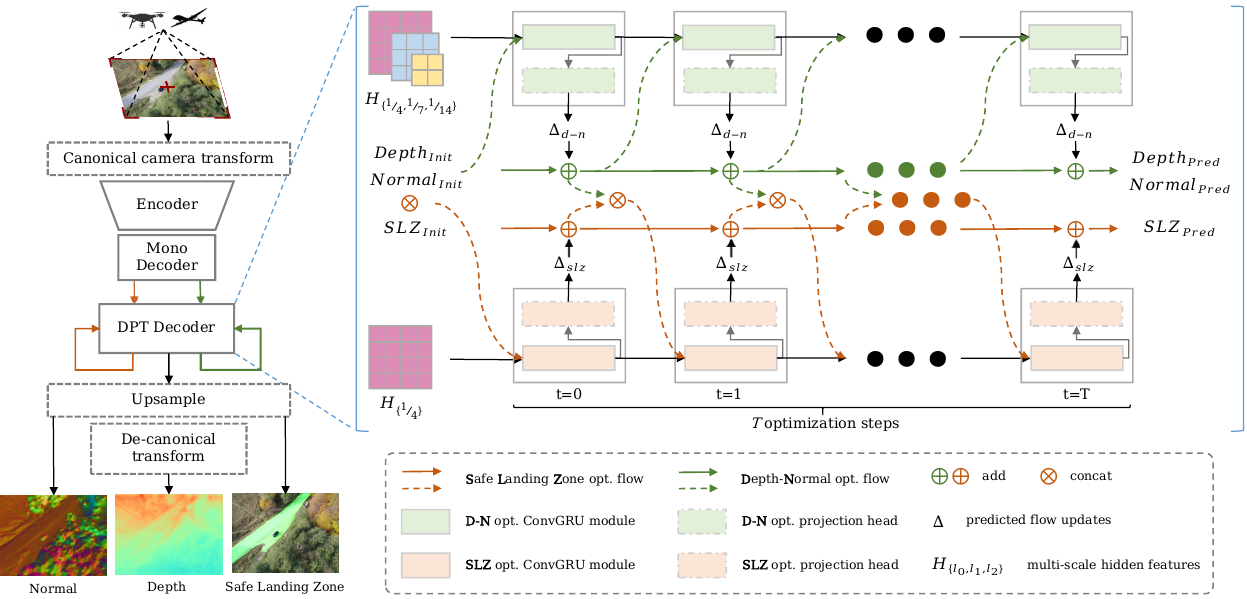}
  \caption{\textbf{Dual-flow optimization framework. (Left) Main pipeline:} Input images are transformed into canonical space via camera parameters $\mathbf{K}$, processed by DINOv2~\cite{dino} ViT~\cite{vit} for multi-scale feature extraction ($F_{1/4}$, $F_{1/7}$, $F_{1/14}$), then refined through $T$ iterations in the Dense Prediction Transformer (DPT~\cite{dpt}). \textbf{(Right) Dual optimization flows:} The depth-normal flow (green) employs three ConvGRU blocks with projection heads for multi-scale refinement $\{H^{1/4}_t, H^{1/7}_t, H^{1/14}_t\}$, while the SLZ flow
  (orange) uses single ConvGRU with geometric fusion of updated depth/normal. Final predictions are obtained through upsampling and inverse canonical transformation.}
  \label{fig2}
\end{figure*}
\section{METHOD}

\subsection{Preliminaries} 

\subsubsection{\textbf{SLZ estimation via segmentation}}
In recent studies based on segmentation approaches, there has been a growing trend toward adopting multi-level risk assessment frameworks, which divide landing zones into \( K \) discrete risk levels (\( K \geq 3 \)). However, such continuous grading strategies can increase the decision-making complexity of the model. Therefore, we argue that it is sufficient to distinguish between safe and unsafe regions, while imposing stricter constraints on the conditions for determining safety to reduce the risk of model misjudgment and avoid potential losses. Based on this, we formulate the SLZ estimation as a binary segmentation problem: given an RGB image \( I \in \mathbb{R}^{H \times W \times 3} \) captured by a drone, a deep neural network \( f_\theta: I \mapsto M \) is used to map it to a binary segmentation mask \( M \in \{0,1\}^{H \times W} \), where:

\[
m_{ij} = 
\begin{cases} 
0 & \text{(safe pixel)} \\
1 & \text{(unsafe pixel)}
\end{cases}
\]

\subsubsection{\textbf{Metric3D V2 model}}
The Metric3D V2 model achieves monocular geometry perception through two key techniques: firstly, it resolves the scale ambiguity of images by employing canonical camera space transformation, and secondly, it introduces a novel depth-normal joint optimization mechanism. The specific pipeline of the model is as follows: First, the input image $I_R$ and depth label $D_R^\text{gt}$ undergo a projection transformation with canonical focal length $f_C$ to generate the canonical space image $D_C$ and its corresponding depth label $D_C^\text{gt}$. Subsequently, the encoder extracts multi-scale features from the image, and the decoder predicts the initial canonical depth $D^0$ and unnormalized normal $N^0$. Finally, these predictions are fed into the DPT~\cite{dpt} module for $T$-step iterative joint optimization, outputting the final predicted canonical depth $D_C$ and predicted normal $N$. During the training process, the model employs canonical space depth labels to supervise depth prediction, while using normal labels to supervise the prediction process of the normal branch. Considering the scarcity of normal label data, Metric3D V2 innovatively proposes to utilize the predicted real depth $D_R$ to impose depth-normal consistency constraints on normal prediction, thereby providing supplementary supervision for normal prediction.

In terms of this task, a qualified SLZ must satisfy three basic conditions: flat terrain, absence of obstacles, and sufficient landing area~\cite{c2}. The depth and normal information predicted by the Metric3D V2 model is closely related to these factors, providing crucial 3D prior information for SLZ estimation and further enabling the estimation of landing zone area. However, existing research still exhibits shortcomings in underutilizing such 3D information.

\subsection{SLZ Estimation via Depth-Normal Joint Optimization}

Autonomous UAVs landing presents a highly challenging task that requires the system to accurately understand the geometric characteristics and safety of the environment. Previous deep learning-based approaches often fail to fully utilize 3D information, tend to produce physically implausible predictions, and typically suffer from overfitting to specific scene data.

\subsubsection{\textbf{Pipeline}}
Building upon Metric3D V2, we propose a novel end-to-end multi-task collaborative optimization framework that achieves comprehensive understanding of SLZ through joint learning of depth estimation, normal prediction, and safe landing zone estimation. The overall pipeline largely follows the design of Metric3D V2 (as shown in the left part of Figure.\ref{fig2}). For an image \( I \in \mathbb{R}^{H \times W \times 3} \) captured by the UAVs, we first transform it to the canonical camera space defined by Metric3D V2 based on camera parameters. In the canonical camera space, we employ DINO Vision Transformer to extract multi-scale image features, followed by a decoder that generates initial depth-normal prediction flow and SLZ prediction flow in low-resolution space. These flows are then iteratively optimized $T$ times through the DPT module. Finally, we obtain predicted depth, normal, and SLZ in real space through upsampling and de-canonical space transformation (Note: only depth prediction requires canonical/real space transformation).

\subsubsection{\textbf{SLZ Estimation Optimization}}
In the Metric3D V2 architecture, the iterative optimization process of DPT only includes a single depth-normal data flow. To fully utilize its pre-trained capabilities while maintaining the original structure, we propose a dual-flow optimization framework as shown in the right part of Figure.\ref{fig2}, which adds a SLZ prediction branch. The core of the DPT module consists of a recurrent update module, which contains ConvGRUs and a projection head: ConvGRUs enhances spatial feature processing capability by introducing convolutional operations, while the projection head (a lightweight CNN) is responsible for predicting the update quantities of various parameters. In the initial optimization stage, DPT receives multi-scale hidden features (containing rich semantic information) output from the decoder. ConvGRUs updates the hidden feature $H$ by integrating all input variables, and after predicting the update quantities through the projection head, it passes the updated hidden features and prediction information to the next iteration step until $T$ optimizations are completed. The two data flows adopt differentiated designs:
\newline\textbf{Depth-Normal Flow} (green part): Contains 3 ConvGRU sub-blocks and 2 projection heads, processing multi-scale feature maps $\{H_t^{1/4}, H_t^{1/7}, H_t^{1/14}\}$, with depth and normal information as input variables. Its optimization process can be described as follows:

\begin{equation}
\begin{array}{l}
    \{H_{t+1}^{1/4}, \cdots \} = \mathrm{ConvGRUs}(D_t \otimes N_t, \{H_t^{1/4}, \cdots \}), \\
    \begin{aligned}[b]
        \Delta D_{t+1} &= \mathrm{Proj_d}(H_{t+1}^{1/4}), & D_{t+1} &= \Delta D_{t+1} + D_{t}, \\
        \Delta N_{t+1} &= \mathrm{Proj_n}(H_{t+1}^{1/4}), & N_{t+1} &= \Delta N_{t+1} + N_{t}
    \end{aligned}
\end{array}
\end{equation}

\textbf{Safe Landing Zone Flow} (orange part): Adopts a single ConvGRU sub-block and projection head design, processing only high-resolution feature maps $H_t^{1/4}$ to control computational costs. During iteration, it integrates SLZ, updated depth, and normal information to collaboratively optimize safe zone prediction through 3D geometric features. The optimization process can be described as follows:
\begin{equation}
\begin{array}{l}
    \mathbb{H}_{t+1}^{1/4} = \mathrm{ConvGRU}(D_{t+1} \otimes N_{t+1} \otimes SLZ_t  , H_t^{1/4}), \\
    \begin{aligned}[b]
        &\Delta SLZ_{t+1} = \mathrm{Proj_{slz}}(\mathbb{H}_{t+1}^{1/4}), \\ &SLZ_{t+1} = \Delta SLZ_{t+1} + SLZ_{t}
    \end{aligned}
\end{array}
\end{equation}
\subsubsection{\textbf{Training}}
Since the training data of Metric3D V2 mainly consists of ground-level and indoor scenes, which significantly differ in scale from UAV perspectives, we adopt a two-stage training strategy: first fine-tuning the base model using UAV aerial data, then freezing the backbone network parameters and training the SLZ prediction branch separately.

Referring to the fine-tuning protocol of Metric3D V2, a combined loss function is used in the fine-tuning stage:
\begin{equation}
L_{\text{fine-tune}} = \lambda_1 \cdot L_{\text{vnl}} + \lambda_2 \cdot \sum_{t=0}^{T} \gamma^{T-t}(L_{\text{L1}}^t + L_{\text{conf}}^t) + \lambda_3 \cdot L_{\text{dncl}}
\end{equation}
where:
\begin{itemize}
    \item The balancing coefficients $\lambda_1$, $\lambda_2$, and $\lambda_3$ control the contribution of each loss term, which are set to $0.2$, $0.5$, and $0.01$, respectively. The temporal decay factor $\gamma$, which controls the weight decay rate of historical predictions, is set to $0.9$. Here, $t \in [0, T]$ denotes the $t$-th prediction stage, where the prediction at $t=0$ refers to the initial prediction provided by the Mono Decoder.
    \item Virtual Normal Loss ($L_{\text{vnl}}$) randomly samples N sets of three points in the image plane to form virtual planes, evaluating the difference between the virtual normals computed from the predicted depth and the ground truth:
    \begin{equation}
    L_{\text{vnl}} = \frac{1}{N}\sum_{i=1}^{N}\|\hat{n}_{\text{pred}}^{(i)} - \hat{n}_{\text{gt}}^{(i)}\|
    \end{equation}
    Here, $\hat{n}_{\text{pred}}$ and $\hat{n}_{\text{gt}}$ denote the unit normal vectors computed from the predicted depth $\hat{D}_{\text{pred}}$ and the ground truth depth $D_{\text{gt}}$, respectively.
    \item Sequential Depth Loss is a temporally optimized term specifically designed for multi-stage prediction networks, encompassing pixel-wise L1 depth regression loss and confidence calibration loss:
    \begin{equation}
        \begin{split}
        &L_{\text{L1}}^t = \frac{1}{N}\sum_{i=1}^{N}|\hat{D}_i^t - D_i^{gt}|,\\ \quad &L_{\text{conf}}^t = \frac{1}{M}\sum_{i=1}^{M}|\hat{C_i}^t - C_i^{gt}|
        \end{split}
    \end{equation}
    Here, $D_i^{\text{gt}}$ denotes the depth label transformed into the canonical camera space, and $C_i^{\text{gt}}$ represents the corresponding confidence label.
    \item Depth-Normal Consistency Loss ($L_{\text{dncl}}$) leverages the camera intrinsic parameters $K$ to establish a differentiable transformation between the depth map and the normal map, enforcing geometric consistency:
    \begin{equation}
    L_{\text{dncl}} = \frac{1}{M}\sum_{i=1}^{M}(1 - \hat{N}_i \cdot N_i)
    \end{equation}
    Here, $\hat{N}_i$ represents the normal vector converted from the predicted depth, and $N_i$ denotes the output of the normal estimation branch. The dot product operation measures the directional consistency between the two.
\end{itemize}

Inspired by the Sequential Depth Loss, we adopt a \textbf{weighted cross-entropy loss} to optimize the SLZ estimation branch. The mathematical formulation of the loss is defined as follows:
\begin{equation}
L_{\text{safe}} = \sum_{t=0}^{T}\gamma^{T-t}\left[-\frac{1}{N}\sum_{j=1}^{N}\sum_{c=1}^{C}Y_{j,c}\log(P_{t,j,c})\right]
\end{equation}
Here, the definitions of $\gamma$, $T$, and $t$ remain consistent with the previous sections, $Y_{j,c}$ represents the ground truth label for the $j$-th pixel and class $c$, and $P_{t,j,c}$ denotes the predicted probability for the $j$-th pixel belonging to class $c$ at time step $t$.Additionally, class-specific weights are incorporated into the loss computation to address class imbalance, with a weight of 2 assigned to the \textit{safe} class and a weight of 1 assigned to the \textit{unsafe} class.
\section{EXPERIMENTS}
\subsection{Datasets}
\subsubsection{\textbf{WildUAV}}
The WildUAV dataset is a real-world image dataset designed for UAV environmental perception tasks, primarily used for validating depth estimation and semantic segmentation algorithms in complex scenarios. The data was captured using a DJI Matrice 210 RTK V2 UAV equipped with a Zenmuse X5s camera system, with image resolutions of 5280$\times$3956 pixels and 3840$\times$2160 pixels. The UAV operated at altitudes ranging from 20 to 30 meters, covering diverse scenes such as fields and forests, and provides precise absolute depth annotations. Due to the lack of safety zone annotations in the original dataset, we manually annotated 1329 images from the mapping subset based on the principle of "flat and obstacle-free" with ISAT\_with\_segment\_anything tool~\cite{ISAT_with_segment_anything}, dividing them into 1065 training samples and 264 validation/test samples. To meet experimental requirements, all images were resized to 1/4 of their original resolution for training and evaluation. Example data are shown in Fig.~\ref{wilduav}.

\subsubsection{\textbf{Semantic Drone}}
\begin{figure}[thpb]
  \centering
  \subfigure[WildUAV]{
    \includegraphics[width=0.25\textwidth]{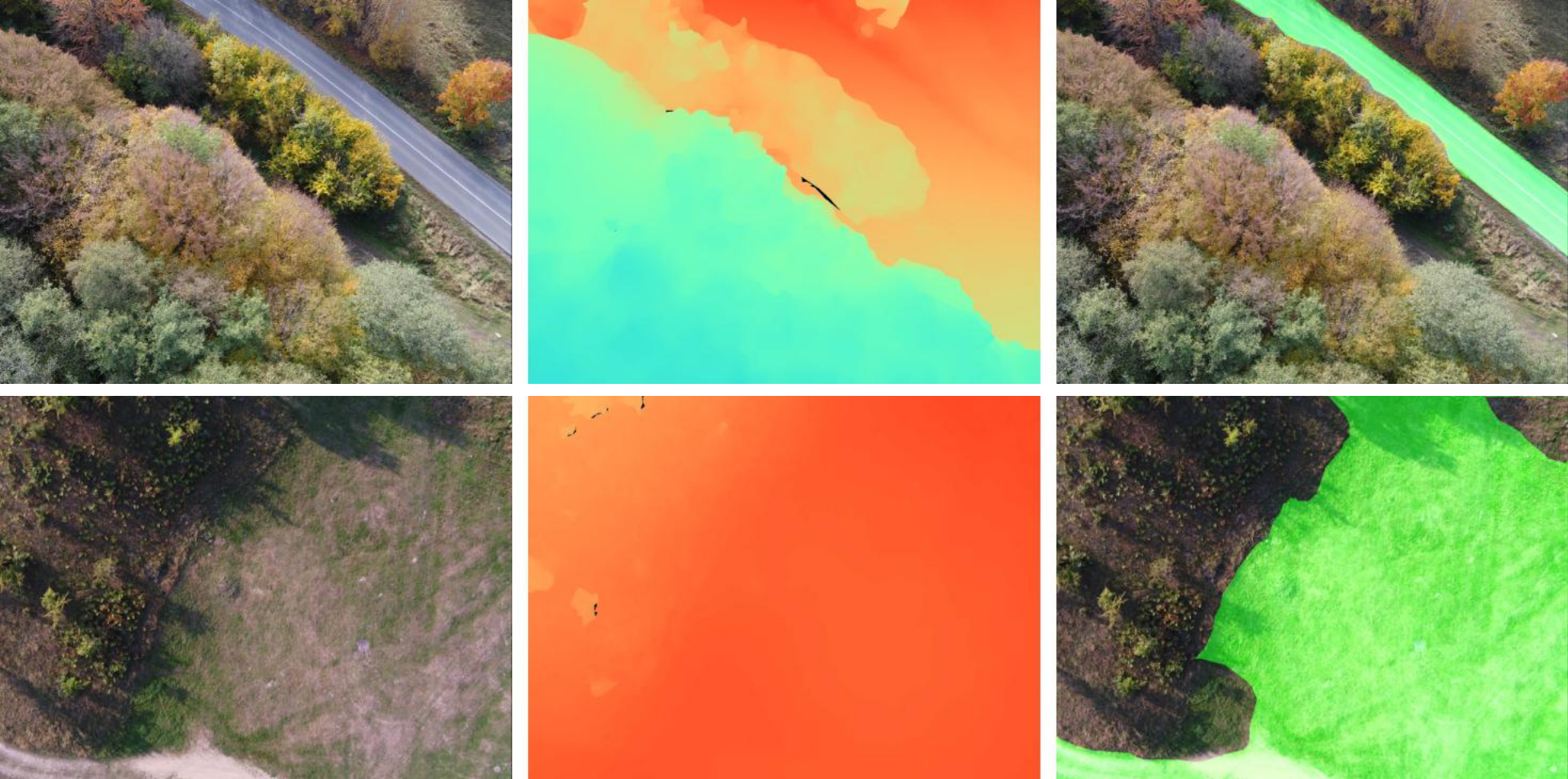}
    \label{wilduav}
  }
  \hfill
  \subfigure[Semantic Drone]{
    \includegraphics[width=0.1845\textwidth]{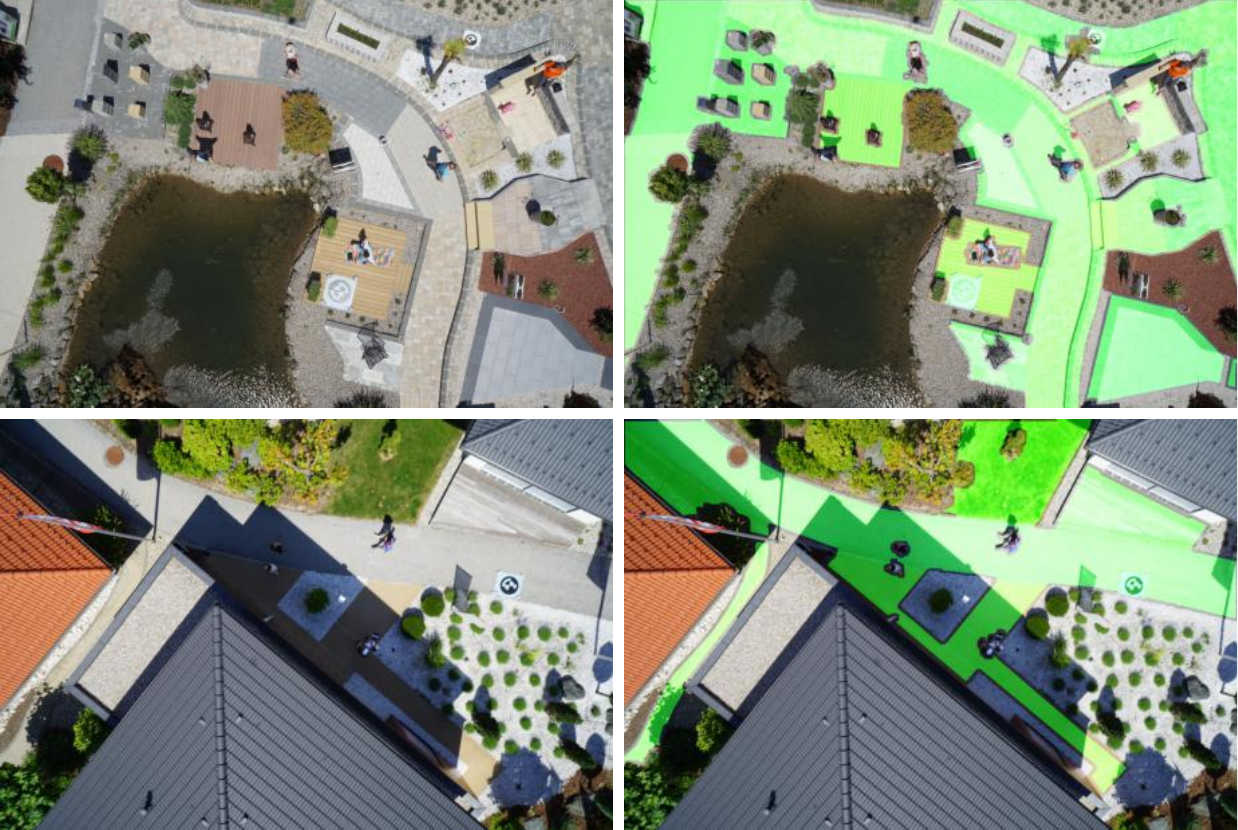}
    \label{semanticdrone}
  }
  \caption{Sample Examples from WildUAV and Semantic Drone Datasets. The WildUAV dataset provides annotations for both depth estimation and safe landing zones (highlighted by green masks), while the Semantic Drone dataset includes annotations exclusively for safe landing zones (indicated by green masks).}
  \label{fig:combined}
\end{figure}

The Semantic Drone dataset focuses on semantic segmentation tasks in urban scenes, aiming to enhance the safety of UAV autonomous flight and landing. The dataset contains high-resolution images of 6000$\times$4000 pixels, captured at altitudes ranging from 5 to 30 meters, with annotations covering over 20 semantic categories such as trees, grass, roads, vehicles, and buildings. To address the lack of safety zone annotations, we followed the criteria in~\cite{c21}, defining grass, predefined markers, and paved areas within ``ideal landing zones'' and ``low-risk zones'' as safety zones, while categorizing all other classes as non-safety zones. Additionally, morphological dilation was applied to the boundaries of non-safety zones to simulate high-risk buffer areas. All 400 publicly available images were used as the cross-domain test set, resized to 1/4 of their original resolution for testing. Example data are shown in Fig.~\ref{semanticdrone}.

\subsection{Implementation Details}
To balance the algorithm's accuracy and real-time requirements, this study improves upon a small version of the Metric3D V2 pretrained model (800k steps of pretraining). Under the premise of adhering to Metric3D V2's fine-tuning protocol, the initial learning rate for the encoder is set to $1 \times 10^{-5}$, while the initial learning rate for the decoder is set to $1 \times 10^{-4}$. The Adam optimizer is employed, with the DPT optimization module set to iterate for $T=4$ steps. In the data augmentation module, the RandomResize ratio range is set to $[0.5, 0.99]$, and the batch size is set to 4, with a total of 20k training steps. To fully utilize the unlabeled data from WildUAV, a semi-supervised training framework is introduced starting from the 4k step, updating the teacher model parameters using the Exponential Moving Average (EMA) strategy. In the second phase, all parameters of the original Metric3D model are fixed, and the safety landing zone prediction branch (SLZ flow) is trained separately, with an initial learning rate of $1 \times 10^{-4}$, a batch size of 4, and a training duration of 5k steps. The experiments are implemented based on the PyTorch and MMSegmentation frameworks, with the hardware platform being an NVIDIA RTX 4080 GPU.

\begin{figure*}[thpb]
  \centering
  \includegraphics[width=0.98\textwidth]{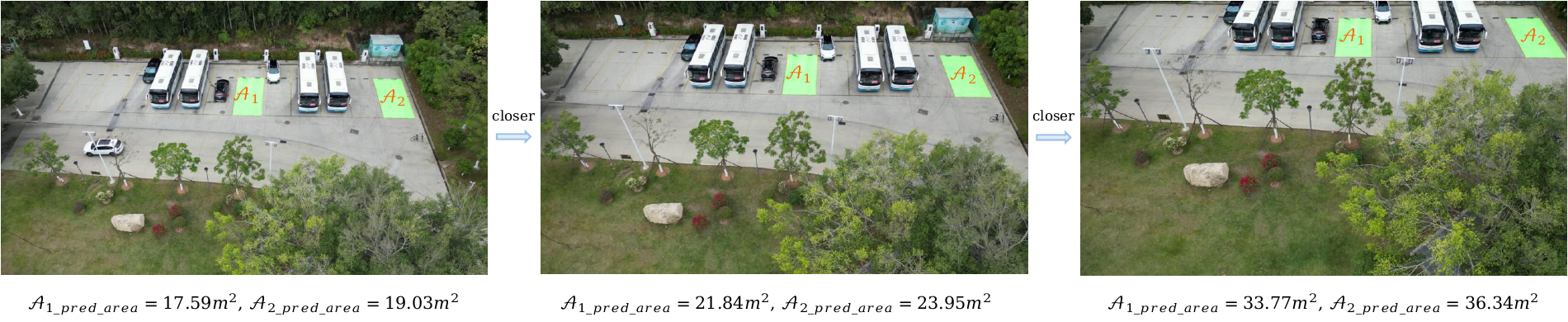}
  \caption{Experimental results of safe landing zone area estimation in real-flight images show improved accuracy when closer to targets, with both measured areas being approximately 34 m².}
  \label{area_fig}
\end{figure*}
\begin{figure*}[thpb]
  \centering
  \includegraphics[width=0.8\textwidth]{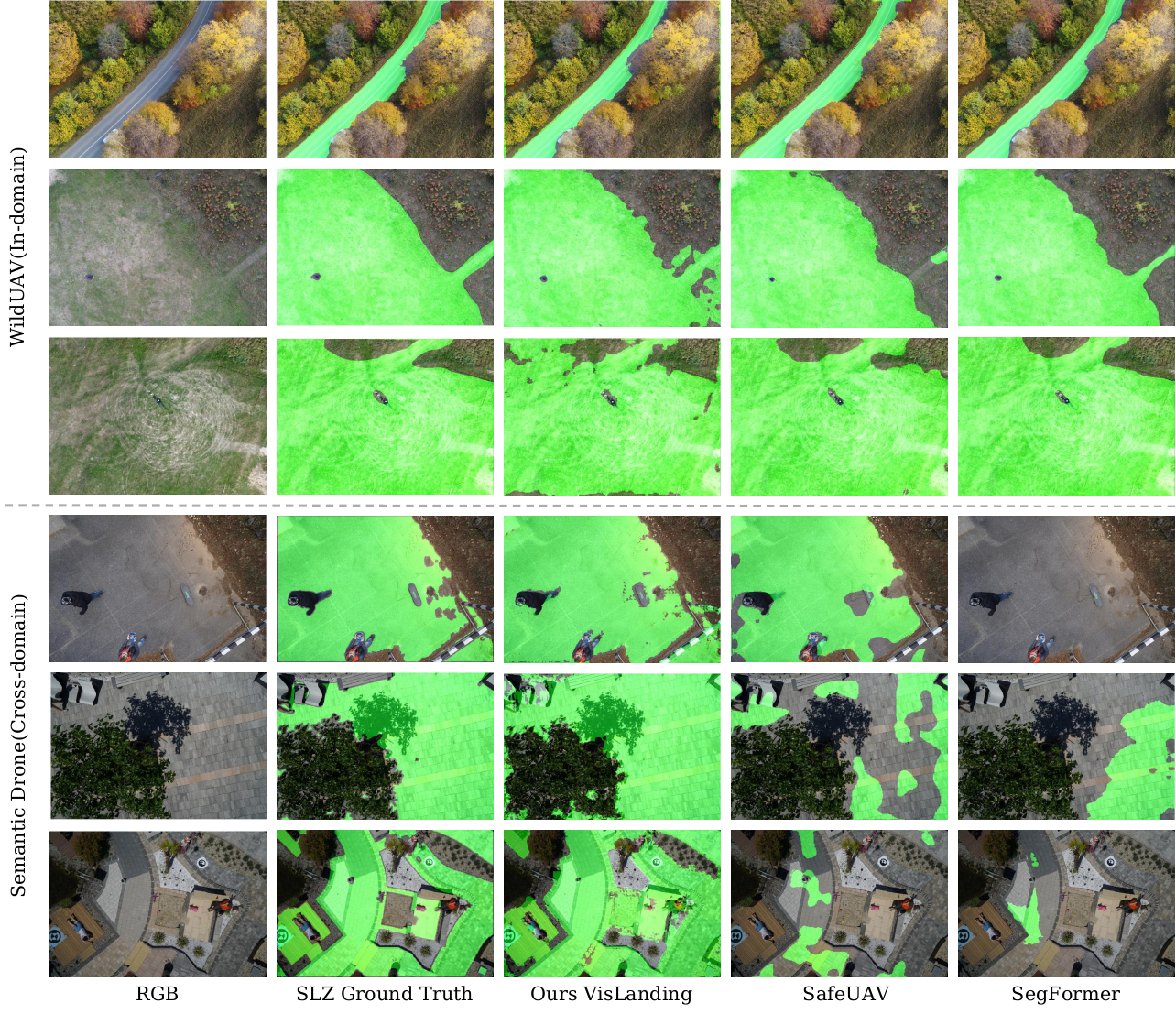}
  \caption{Qualitative prediction results of our method compared with other approaches on two test benchmarks. The proposed method demonstrates significant advantages in terms of generalization capability and robustness, outperforming other methods in challenging scenarios.}
  \label{pre_vis}
\end{figure*}

\subsection{Evaluation and Comparison}
\begin{table*}[!h]
\centering
\caption{Model performance comparison results on in-domain and cross-domain metrics,\\ \textbf{bold} and \underline{underline} denote the best and second-best results.}
\label{tab:model_comparison}
{\fontsize{7.2}{10}\selectfont
\setlength{\tabcolsep}{2.8pt}
\begin{tabular}{@{}l*{14}{c}@{}}
\toprule
\multirow{2}{*}{\textbf{Methods}} 
& \multicolumn{7}{c}{\textbf{In-Domain}} & \multicolumn{7}{c}{\textbf{Cross-Domain}} \\
\cmidrule(lr){2-8} \cmidrule(l){9-15}
& aAcc$\uparrow$ & mIoU$\uparrow$ & mAcc$\uparrow$ & mDice$\uparrow$ & mFscore$\uparrow$ & mPrecision$\uparrow$ & mRecall$\uparrow$ 
& aAcc$\uparrow$ & mIoU$\uparrow$ & mAcc$\uparrow$ & mDice$\uparrow$ & mFscore$\uparrow$ & mPrecision$\uparrow$ & mRecall$\uparrow$ \\
\midrule
SafeUAV\_big           & 94.16 & 84.01 & 89.97 & 91.06 & 91.06 & 92.28 & 89.97 & 59.33 & 39.61 & 61.72 & 55.59 & 55.59 & \textbf{70.27} & 61.72 \\
SafeUAV\_big\_pretrain   & 96.03 & 88.66 & 92.32 & 93.86 & 93.86 & 95.65 & 92.32 & 52.15 & 30.58 & 55.18 & 43.86 & 43.86 & 67.50 & 55.18 \\
segformer-mit-b0       & \textbf{97.78} & \textbf{93.72} & \textbf{97.12} & \textbf{96.72} & \textbf{96.72} & \underline{96.34} & \textbf{97.12} & 53.12 & 31.99 & 56.00 & 45.97 & 45.97 & 66.50 & 56.00 \\
segformer-mit-b1       & \underline{97.61} & \underline{93.12} & \underline{95.87} & \underline{96.39} & \underline{96.39} & \textbf{96.94} & \underline{95.87} & 55.82 & 35.78 & 58.33 & 51.18 & 51.18 & 65.74 & 58.33 \\
\midrule
Ours ($T$=1) & 90.44 & 75.56 & 84.70 & 85.45 & 85.45 & 86.26 & 84.70 & 65.57 & 48.65 & 65.43 & 65.42 & 65.42 & 65.41 & 65.43 \\
Ours ($T$=2) & 90.90 & 76.41 & 84.99 & 86.03 & 86.03 & 87.22 & 84.99 & 68.52 & 52.08 & 68.61 & 68.48 & 68.48 & 68.52 & 68.61 \\
Ours ($T$=3) & 91.05 & 76.31 & 84.15 & 85.93 & 85.93 & 88.18 & 84.15 & \underline{68.54} & \underline{52.13} & \underline{68.99} & \underline{68.53} & \underline{68.53} & 69.09 & \underline{68.99} \\
Ours ($T$=4) & 91.15 & 76.61 & 84.46 & 86.15 & 86.15 & 88.23 & 84.46 & \textbf{68.81} & \textbf{52.44} & \textbf{69.35} & \textbf{68.79} & \textbf{68.79} & \underline{69.55} & \textbf{69.35} \\
\bottomrule
\end{tabular}}
\end{table*}
In this study, centered around the practical requirements of SLZ segmentation tasks, we designed a dual-dimensional evaluation system encompassing in-domain and cross-domain assessments. While ensuring fundamental segmentation accuracy, the system primarily validates the model's domain generalization capability and robustness. The experiments covered four models from the SafeUAV series (big/big-pretrain) and SegFormer (mitb0/mitb1), with evaluation metrics including:
\begin{itemize}
    \item \textbf{aAcc (Average Accuracy)}: The average classification accuracy across all classes.
    \item \textbf{mIoU (Mean Intersection over Union)}: The average IoU across all classes, measuring the overlap between predicted and ground truth regions.
    \item \textbf{mAcc (Mean Accuracy)}: The average pixel accuracy across all classes.
    \item \textbf{mDice (Mean Dice Coefficient)}: The average Dice coefficient across all classes, evaluating the overlap between predicted and ground truth regions.
    \item \textbf{mFscore (Mean F1 Score)}: The average F1 score across all classes, balancing Precision and Recall.
    \item \textbf{mPrecision (Mean Precision)}: The average Precision across all classes, measuring the accuracy of positive predictions.
    \item \textbf{mRecall (Mean Recall)}: The average Recall across all classes, measuring the completeness of positive predictions.
\end{itemize}

The training settings strictly adhered to benchmark methods: SegFormer was fine-tuned for 30 epochs based on ImageNet-1k pre-trained weights~\cite{c21}, while the base SafeUAV models were trained on WildUAV for 50 epochs~\cite{safeuav}. The pre-trained versions were first pre-trained on the HOV dataset~\cite{safeuav} for 50 epochs and then transferred to WildUAV for further training. Our evaluation experiments are conducted in two dimensions:
\begin{itemize}
    \item \textbf{In-domain}: Models are trained and tested on the WildUAV dataset.
    \item \textbf{Cross-domain}: Models are trained on the WildUAV dataset but directly transferred to the Semantic Drone dataset for testing.
\end{itemize}
The WildUAV dataset consists of images captured in wild environments, while the Semantic Drone dataset contains images collected in urban settings. The significant domain gap between these two datasets makes them particularly suitable for cross-domain evaluation, effectively assessing the generalization capability and robustness of the models.

As shown in the left part of Table~\ref{tab:model_comparison}, all models achieved usable performance levels in the in-domain tests based on the aAcc (Average Accuracy) metric. Among them, SegFormer demonstrated significant superiority across all metrics due to its large-scale pre-training advantage on segmentation tasks. In contrast, our method, which did not undergo task-specific pre-training for segmentation, exhibited a measurable performance gap. However, in the cross-domain tests (right part of Table~\ref{tab:model_comparison}), the benchmark models experienced a notable performance degradation. Taking the key metrics aAcc and mIoU as examples, the SegFormer series showed a decline of approximately 45\% and 64\%, respectively, compared to its in-domain performance, while the SafeUAV series dropped by about 42\% and 60\%. In stark contrast, our method only decreased by approximately 25\% and 33\%, and significantly outperformed the benchmark models in absolute performance metrics. This performance disparity is further corroborated by the qualitative comparisons in Fig.~\ref{pre_vis}, where our method maintains consistent prediction coherence under domain shifts while competitors exhibit fragmented outputs. This result indicates that traditional semantic segmentation methods suffer from clear domain-specific biases, leading to severe overfitting issues. In comparison, our method effectively mitigates cross-domain performance degradation by leveraging 3D information, demonstrating stronger applicability in real-world complex environments.

Furthermore, to address real-time requirements, we conducted experiments with different optimization iteration steps, as shown in the lower part of Table~\ref{tab:model_comparison}. The results indicate that even with only 1 optimization step, the performance degradation remains entirely acceptable, with minimal impact on the model's generalization capability, and our method still maintains a significant advantage in cross-domain tests. This characteristic enables our method to significantly improve inference speed in computationally constrained scenarios while maintaining high performance levels.

\subsection{Estimation of Safe Landing Zone Area}
Our model also inherits the powerful zero-shot prediction capabilities of Metric3D V2 for absolute depth and normal maps. By leveraging this information along with the intrinsic matrix of the onboard camera, we can achieve end-to-end rapid rough estimation of safe zone areas, providing critical information for landing site selection. Using the camera intrinsic matrix, depth map, and normal map, pixels in the image are projected into the real-world coordinate system, and the tilt effect is corrected using normal information to accurately estimate the actual area of the selected polygonal region. Specifically, given the camera intrinsic matrix $ K $ and depth map $ D $, the pixel coordinates $ (u,v) $ in the image can be projected to the 3D point $ (X_c,Y_c,Z_c) $ in the camera coordinate system, calculated as:
\begin{equation}
\begin{cases}
X_c = \dfrac{(u - c_x) \cdot D(u, v)}{f_x} \\
Y_c = \dfrac{(v - c_y) \cdot D(u, v)}{f_y} \\
Z_c = D(u, v)
\end{cases}
\end{equation}
where $ f_x $ and $ f_y $ are the focal lengths of the camera, and $ c_x $ and $ c_y $ are the principal point coordinates. The surface tilt effect is corrected using the Z component of the normal vector $ \mathbf{n} = (n_x, n_y, n_z) $, and the projected area of a single pixel is calculated as:
\begin{equation}
A_{\mathrm{actual}}(u, v) = { \dfrac{D(u, v)^2}{{f_x \cdot f_y}\cdot{|n_z(u, v)|}}}
\end{equation}
The total actual area is obtained by summing the actual area contributions of all pixels within the selected polygonal region:
\begin{equation}
A_{\mathrm{total}}=\sum_{(u, v) \in \mathrm{SLZ}} A_{\mathrm{actual}}(u, v)
\end{equation}
In tests on real flight data not involved in training (as shown in Figure.\ref{area_fig}), estimation experiments on two preset safe landing zones with areas around 34m² showed that as the drone approached the target area, the improvement in depth estimation accuracy significantly reduced area errors. Notably, the system exhibited conservative estimation characteristics in long-distance scenarios (estimates systematically lower than actual measurements), providing safety redundancy for autonomous landing decisions.
\subsection{Limitations}
Although our method has already demonstrated good generalization performance and low hardware complexity in real-world scenarios, its performance has not been validated on larger-scale datasets. Therefore, we plan to collect and construct larger datasets for training and evaluation in the future.
Despite the relatively lower in-domain metrics compared to other methods, overfitting on in-domain data often leads to reduced generalization on cross-domain tasks, which is a common issue in the field. Thus, we prioritize generalization performance in real-world scenarios over merely achieving good results on test sets with distributions identical to the training set. Furthermore, in subsequent work, we will further investigate methods to achieve strong performance in both in-domain and cross-domain settings. However, due to the large parameter size and high inference cost of the Vision Transformer model, our method cannot meet the high real-time requirements (e.g., $<$5 FPS on Jetson Orin Nano) for practical applications such as online processing on drones. To address this, we will explore techniques such as model distillation and transfer learning to enhance the practical potential of our approach.

\section{CONCLUSIONS}
The proposed \textbf{VisLanding} framework achieves precise estimation of SLZ for UAVs under monocular vision through a depth-normal collaborative optimization mechanism. Experimental results demonstrate that the proposed method attains a 76.61\% mIoU for SLZ segmentation on the WildUAV dataset and maintains a 52.44\% mIoU in cross-domain testing (on the Semantic Drone dataset), significantly outperforming traditional semantic segmentation methods. By leveraging depth and normal information, the system can estimate the actual area of landing zones in real time, providing critical support for decision-making. Compared to multi-sensor fusion approaches, our method significantly reduces hardware complexity while maintaining high accuracy, offering a reliable solution for autonomous UAV landing in emergency scenarios. Future work will focus on exploring online optimization strategies for dynamic complex scenes and further compressing the computational cost of the model to meet real-time requirements.






\end{document}